\documentclass[letterpaper]{article} 
\usepackage{aaai24}  
\usepackage{times}  
\usepackage{helvet}  
\usepackage{courier}  
\usepackage[hyphens]{url}  
\usepackage{graphicx} 
\usepackage{natbib}  
\usepackage{caption} 
\usepackage{algorithm}
\usepackage{algorithmic}
\usepackage{graphicx}
\usepackage{amsmath}
\usepackage{amsthm}
\usepackage{booktabs}
\usepackage{algorithm}
\usepackage{algorithmic}
\usepackage[switch]{lineno}
\usepackage[marginal]{footmisc}
\usepackage{color} 

\usepackage{multirow}
\usepackage{subfigure}
\usepackage{enumitem}
\usepackage{amsthm,amsmath,amssymb}
\usepackage{mathrsfs}
\usepackage{booktabs}
\usepackage{tabularx}

\newtheorem{myDef}{Challenge}
\newtheorem{myDef1}{Definition}

\usepackage{newfloat}
\usepackage{listings}
\DeclareCaptionStyle{ruled}{labelfont=normalfont,labelsep=colon,strut=off} 
\floatstyle{ruled}
\newfloat{listing}{tb}{lst}{}
\floatname{listing}{Listing}
\title{Exploring Large Language Model for Graph Data Understanding \\in Online Job Recommendations}
\author{
    Likang Wu,\textsuperscript{\rm 1,\rm 2}
    Zhaopeng Qiu,\textsuperscript{\rm 2}
    Zhi Zheng,\textsuperscript{\rm 1,\rm 2}
    Hengshu Zhu,\textsuperscript{\rm 2 $\ast$}
    Enhong Chen,\textsuperscript{\rm 1 $\ast$}
}
\affiliations{
    \textsuperscript{\rm 1} University of Science and Technology of China \\
    \textsuperscript{\rm 2} Career Science Lab, BOSS Zhipin \\
\{wulk,zhengzhi97\}@mail.ustc.edu.cn,
\{zhpengqiu,zhuhengshu\}@gmail.com,
cheneh@ustc.edu.cn
}

\usepackage{bibentry}

\begin{document}

\maketitle
\renewcommand{\thefootnote}{\fnsymbol{footnote}} 
\footnotetext[1]{Corresponding Author.} 
\footnotetext[2]{During the BOSS Zhipin internship period.} 

\begin{abstract}
Large Language Models (LLMs) have revolutionized natural language processing tasks, demonstrating their exceptional capabilities in various domains. However, their potential for graph semantic mining in job recommendations remains largely unexplored. This paper focuses on unveiling the capability of large language models in understanding behavior graphs and leveraging this understanding to enhance recommendations in online recruitment, including promoting out-of-distribution (OOD) applications. We present a novel framework that harnesses the rich contextual information and semantic representations provided by large language models to analyze behavior graphs and uncover underlying patterns and relationships. Specifically, we propose a meta-path prompt constructor that aids LLM recommender in grasping the semantics of behavior graphs for the first time and design a corresponding path augmentation module to alleviate the prompt bias introduced by path-based sequence input. By facilitating this capability, our framework enables personalized and accurate job recommendations for individual users. We evaluate the effectiveness of our approach on comprehensive real-world datasets and demonstrate its ability to improve the relevance and quality of recommended results. This research not only sheds light on the untapped potential of large language models but also provides valuable insights for developing advanced recommendation systems in the recruitment market. The findings contribute to the growing field of natural language processing and offer practical implications for enhancing job search experiences. We release the code\footnotemark[3]. 
\end{abstract}

\section{Introduction}
Online recruitment recommendations aim to suggest relevant job opportunities to job seekers based on their preferences and qualifications, improving the chances of matching the right employment. With the exponential growth of online recruitment platforms and the need for efficient and personalized job search experiences, the development of effective job recommendation systems has become crucial. 

In online recruitment systems, job postings and resumes are written in natural language. Traditional approaches have treated job-resume matching as a supervised text-matching problem using paired data for training~\cite{qin2018enhancing,shen2018joint}. However, online recruitment platforms often suffer from sparse interaction data, with job postings attracting only a few candidates on average~\cite{ramanath2018towards}. To address this, recent studies~\cite{bian2020learning,Yang_Hou_Song_Zhang_Wen_Zhao_2022} have explored the use of behavior graphs to capture high-order interactions and alleviate the sparse interaction issue. These behavior graphs leverage message passing to enhance the understanding of user preferences.

Unlike many general recommendation tasks, it is easy to find that textual understanding forms the backbone of job recommendation, and behavior modeling contributes to the personalized module. In our work, we strive to overcome the accuracy limitations of job recommenders by enhancing the semantic richness of textual representations. Inspired by several recent successful recommendations based on text pre-training~\cite{wu2023survey},  we introduce a large language model (LLM) as the foundational framework for job recommendation that directly generates targets. Adopting this approach is not only beneficial but also intuitive. For instance, out-of-distribution items usually appear in recruitment markets since new job demands are constantly emerging, such as prompt engineers for generative models. This issue is more complex than traditional cross-domain tasks~\cite{zhao2023cross,jiang2023knowledge,yu2023untargeted}. The powerful semantic mining ability and extensive external knowledge of LLMs augment the generation and associative power of recommenders, which is able to generate reasonable recommendation results for the hard OOD items. 

However, the existing learning schema of LLM recommender cannot understand the non-textual behavior graph which weakens the personalized recommendation ability for different job seekers. To tackle this challenge, we propose a meta-path prompt constructor to encode the interaction information of graph into the natural language prompt. Specifically, in such a heterogeneous behavior graph, each meta-path composed of various types of nodes and edges can be transferred into a description naturally since each type indicates a specific and meaningful interaction, e.g., interview, conversation, etc. Along this line, for each job seeker, the LLM captures the high-order interaction feature to augment her personality with the meta-path prompt.

\footnotetext[3]{{https://github.com/WLiK/GLRec}}
Based on the above analysis, we explore the inclusion of graph data understanding in large language model-based recommendations for the first time. An efficient large language model named GLRec (Graph-understanding LLM Recommender) is proposed to optimize the recommended quality of job recommendation, which is fine-tuned with LoRa~\cite{hu2021lora} on our constructed instruction dataset for aligning the gap between pre-trained knowledge and actual recruitment domain. Especially, our exploration presents two valuable and important findings that largely influence the graph understanding strategy of LLM: (i). Different paths would present different weights for the model decision. (ii). The position bias of the order of path prompts brings unstable answers. For these issues, we carefully design path shuffling, adaptive path selector, and their hybrid path augmentation mechanism to mitigate the adverse effects posed by varying path prompts. 
The main contributions could be summarized as follows:
\begin{itemize}
\item To our best knowledge, we are the first to implement the fine-tuned large language model as job recommender, which promotes matching accuracy via the semantic richness and massive knowledge of LLM.
\item We propose the meta-path prompt constructor that leverages LLM recommender to comprehend behavior graphs for the first time and design a corresponding path augmentation module to alleviate the prompt bias.
\item We conduct sufficient experiments on real-world recruitment datasets, and the experimental results and visualization cases show the superiority of our model.
\end{itemize}

\begin{figure*}[t]
    \centering
    \includegraphics[width=0.92\textwidth]{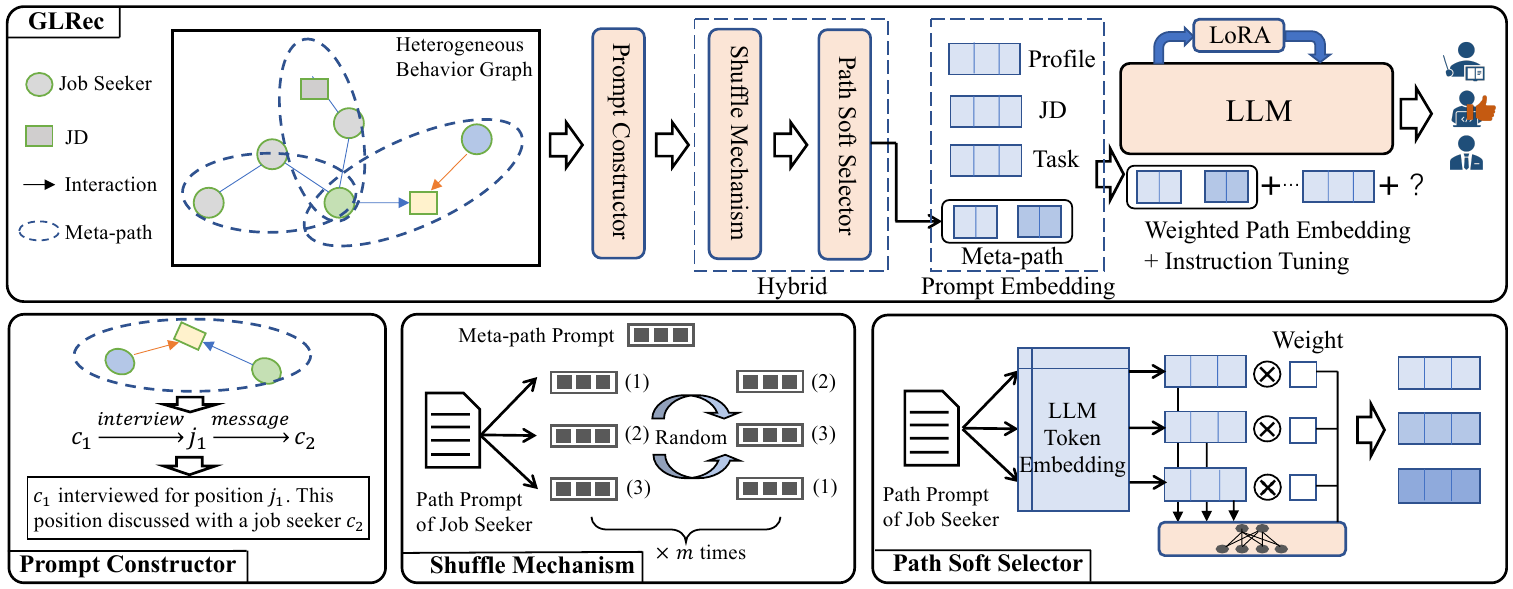}
    \caption{The framework of GLRec for job recommendation.}
    \label{fig:framework}
\end{figure*}
\section{Related Work}
\subsection{Job Recommendation}
Job Recommendation, especially job-resume matching is a necessary task in online recruitment, and it has been extensively studied in the literature~\cite{Kenthapadi_Le_Venkataraman_2017}. Early methods handled this problem~\cite{Lu_El_Helou_Gillet_2013} relying on collaborative filtering assumptions. However, recent research focused more on text-matching technology.
Various techniques have been proposed to encode job and resume information. For example, ~\cite{shen2018joint} utilized CNN for encoding, while~\cite{qin2018enhancing} leveraged RNN and BiLSTM to capture sequential information. ~\cite{Yan_Le_Song_Zhang_Zhang_Zhao_2019} introduced a profiling memory to learn latent preference representation by interacting with both job and resume. ~\cite{luo2019resumegan} explored the effectiveness of adversarial training for job-resume matching. In addition to the aforementioned research, there were also researches that considered multi-granularity interactions. The ranking-based loss can be used to capture multi-level interactions as supervision signals~\cite{le2019towards}. \cite{fu2021beyond} proposed a bilateral multi-behavior sequence model to describe users' dynamic preferences. These approaches highlighted the importance of considering various interaction patterns and incorporating additional user information to improve the quality of job recommendations. However, online recruitment platforms frequently encounter challenges due to sparse interaction data, resulting in job postings attracting only a limited number of candidates on average~\cite{ramanath2018towards}. Recent studies~\cite{bian2020learning,Yang_Hou_Song_Zhang_Wen_Zhao_2022} have investigated the utilization of behavior graphs to capture high-order interactions and alleviate the problem of sparse interactions. 
\subsection{Large Language Models for Recommendation}
LLMs offer the potential to extract high-quality representations of textual features and leverage extensive external knowledge to enhance recommendation systems. ~\cite{wu2023survey} conducted a systematic review and analysis of existing LLM-based recommendation systems. Existing work can be divided into two categories: discriminative models and generative models. Most discriminative models align the representations of pre-trained models like BERT with domain-specific data through fine-tuning.  For example,~\cite{qiu2021u,wu2021userbert} proposed pre-training and fine-tuning-based approach to learn users' representation, which leveraged content-rich domains to complement those users' features with insufficient behavior data. Additionally, some research explored training strategies like prompt tuning. ~\cite{penha2020does} leveraged BERT's Masked Language Modeling (MLM) head to uncover its understanding of item genres using cloze-style prompts. Prompt4NR~\cite{zhang2023prompt} pioneered the application of the prompt learning paradigm for news recommendation. Generative models usually translate recommendation tasks as natural language tasks, and then apply techniques such as in-context learning~\cite{DBLP:journals/corr/abs-2305-08845,DBLP:journals/corr/abs-2212-10559}, prompt tuning~\cite{DBLP:journals/corr/abs-2305-06474,DBLP:journals/corr/abs-2305-00447}, and instruction tuning~\cite{DBLP:journals/corr/abs-2305-07001,DBLP:journals/corr/abs-2205-08084} to adapt LLMs to directly generate the recommendation results. Compared to discriminative models, generative models have better natural language generation capabilities. In the recruitment area, there was a generative model which developed LLM with RLHF to generate potential JDs for more explainable recommendations~\cite{zheng2023generative}. However, despite their successes, LLM recommenders have a glaring limitation: they lack the ability to comprehend graph data, which impedes their potential for personalized adaptation.

\section{Methodology}
In this section, the technical detail of GLRec in Figure~\ref{fig:framework} would be introduced progressively.

\subsection{Preliminary}
\subsubsection{Problem Formulation}
\label{sec:PF}
Consider a set of candidates $C = \{c_1, c_2, \ldots, c_{n_1}\}$ and a set of jobs $\mathcal{J} = \{j_1, j_2, \ldots, j_{n_2}\}$, where $n_1$ and $n_2$ represent the total number of candidates (job seekers) and jobs, respectively. Each candidate and job are associated with textual documents that describe their resumes and job requirements. They are also linked to a collection of directed interaction records (such as interviewing and discussing) within the recruitment platform. These interactions are formally represented as $\mathcal{A}{c_i} = \{c_i \rightarrow j' | c_i \in C, j' \in \mathcal{J}\}$ and $\mathcal{A}{j_k} = \{j_k \rightarrow c' | j_k \in \mathcal{J}, c' \in C\}$, indicating the directed interactions initiated by candidate $c_i$ or employer $j_k$ (referred to as a job). 
Our objective is to predict the compatibility between a job posting and a candidate.
\subsubsection{Generative Large Language Models}
Generative LLMs are powerful language models capable of generating coherent and contextually relevant text. Models like GPT-3 and GPT-4 are trained on vast amounts of text data, enabling them to produce human-like text in response to a given prompt or input. Fine-tuning is a common adaption strategy to align the target of pre-trained model and domain-specific applications, such as two popular paradigms of prompt tuning, and instruction tuning. For all these tuning methods, they have an equal final objective loss of autoregressive training as follows:
{\setlength\abovedisplayskip{0.05cm}
\setlength\belowdisplayskip{0.05cm}
\begin{equation}
\mathcal{L}_{f} = \max _{\Theta} \sum_{(x, y) \in \mathcal{T}} \sum_{t=1}^{|y|} \log \left(\mathcal{P}_{\Theta}\left(y_{t} \mid x, y_{<t}\right)\right),
\label{eq:autog}
\end{equation}
}
Taking instruction tuning as an example, which designs and constructs instruction data to restrict the output scope and format. $x$ and $y$ represent the ``Instruction Input'' and ``Instruction Output'' in the self-instruct data, respectively, e.g., \emph{Instruction Input: ``Do you like this item?'', Instruction Output: ``Yes.''}. And $y_{t}$  is the  $t$-th token of the  $y$, $y_{<t}$  represents the tokens before  $y_{t}$, $\Theta$  is the original parameters of LLM, and  $\mathcal{T}$  is the training set.
\subsubsection{Task-specific Instruction}
In our work, we design two job recommendation tasks to test the LLM recommender following existing related work~\cite{DBLP:journals/corr/abs-2305-00447}, i.e., point-wise and pair-wise job matching. Here we introduce our designed template for the sample in our dataset, where information related to privacy and business has been filtered. Assume there is a job seeker called candidate whose Candidate Profile Prompt and recommended JD Prompt are defined as:

\setlength{\fboxrule}{0.8pt}
\noindent\fbox{\parbox{\linewidth}{
\noindent\textbf{Candidate Profile Prompt:} \emph{Age: 25, Education: Bachelor's degree, Graduation School: XXX
University, Major: Computer Applied Science, Work Experience: 2 years}.

\noindent\textbf{JD Prompt:} \emph{Position Title: Full Stack Engineer, Educational Requirement: Bachelor's degree, Work Experience: 1-3 years, Skill Requirements: HTML/JAVA/Spring Boot/SQL.}
}}

\noindent For the point-wise task, we let the LLM recommender learn to predict the satisfaction of a candidate with a recommended job. The instruction is designed as:

\noindent\fbox{\parbox{\linewidth}{
\noindent\textbf{Point-wise Instruction:} \emph{You are a recommender, determining whether a candidate would be satisfied with the recommended job position. Please answer with ``Yes." or ``No."}.
}}

\noindent For the pair-wise task, we let the LLM recommender learn to justify the preference of a candidate for a recommended job pair. Given two jobs' JD Prompt ``A" and ``B", the instruction is designed as:

\noindent\fbox{\parbox{\linewidth}{
\noindent\textbf{Pair-wise Instruction:} \emph{You are a recommender, determining which position will match the candidate. Please answer with ``[A]." or ``[B]."}.
}}

With the above-designed prompts and instruction, LLM is able to adapt to a domain recommendation situation. Note that, to ensure the stability of training, we append the JD prompt to the end of the ground truth to increase the predicted length. To further fuse interaction knowledge, in the next section, we will illustrate the understanding part of graph data for LLM: behavior meta-path prompt generation.

\subsection{Behavior Meta-path Prompt Generation}
To equip LLM with the ability to comprehend interactive relationships in graph data, we propose a meta-path-based prompt constructor to obtain prompt inputs that represent local subgraphs. Before delving into the details of our approach, it is necessary to provide a formal introduction to heterogeneous graph and meta-path~\cite{wu2021learning}.

\begin{myDef1}
\textbf{ Heterogeneous  Graph. }
	$\mathcal{G}=(V, E)$, consists of an object set $V$ and a link set $E$.
	$\mathcal{G}$ is also associated with a node type mapping function $\phi: V \rightarrow \mathcal{V}$ and 
	a link type mapping function $\psi: E \rightarrow \mathcal{E}$. $\mathcal{V}$ and $\mathcal{E}$ denote 
	the sets of predefined object types and link types, where $|\mathcal{V}|+|\mathcal{E}|>2$.
\end{myDef1}

\begin{myDef1}
\textbf{ Meta-path. }
	A meta-path $P$ is defined as a path in the form of $\mathcal{V}_1 \xrightarrow{\mathcal{E}_1} \mathcal{V}_2 \xrightarrow{\mathcal{E}_2} \cdots \xrightarrow{\mathcal{E}_l} \mathcal{V}_{l+1}$ (abbreviated as $\mathcal{V}_1 \mathcal{V}_2 \cdots \mathcal{V}_{l+1}$), which describes a composite relation $\mathcal{E}_1 \circ \mathcal{E}_2 \circ \cdots \circ \mathcal{E}_l$ between objects $\mathcal{V}_1$ and $\mathcal{V}_{l+1}$, where $\circ$ denotes the composition operator on relations.
\end{myDef1}

Heterogeneous graphs are more diverse and complex in terms of their semantics compared to homogeneous graphs. Meta-paths are commonly used techniques to mine and represent the interaction semantics within them. In the context of online recruitment, the interactions between job seekers and job positions, which involve different types of behaviors, form a behavior graph. This behavior graph is a typical heterogeneous graph, where different node types include {Candidate, JD}, and different edge types include messaging, interviewing, matching, and more.

Due to the unique and defined semantics of each type of edge in the behavior graph, it is natural to consider transferring the graph data format meta-path to a natural language description which is acceptable for the large language model. We only need to predefine the prompt template according to the appeared edges in a path and then fill in the template with the resume or job description information. For instance, given a typical meta-path $c_1 \xrightarrow{interview} j_1 \xrightarrow{message} c_2$. The prompt template is constructed as: 

\noindent \textbf{Meta-path Prompt:} \emph{$c_1$ interviewed for position $j_1$. This position discussed with a job seeker $c_2$}.

\begin{figure}[t]
    \centering
    \includegraphics[width=0.48\textwidth]{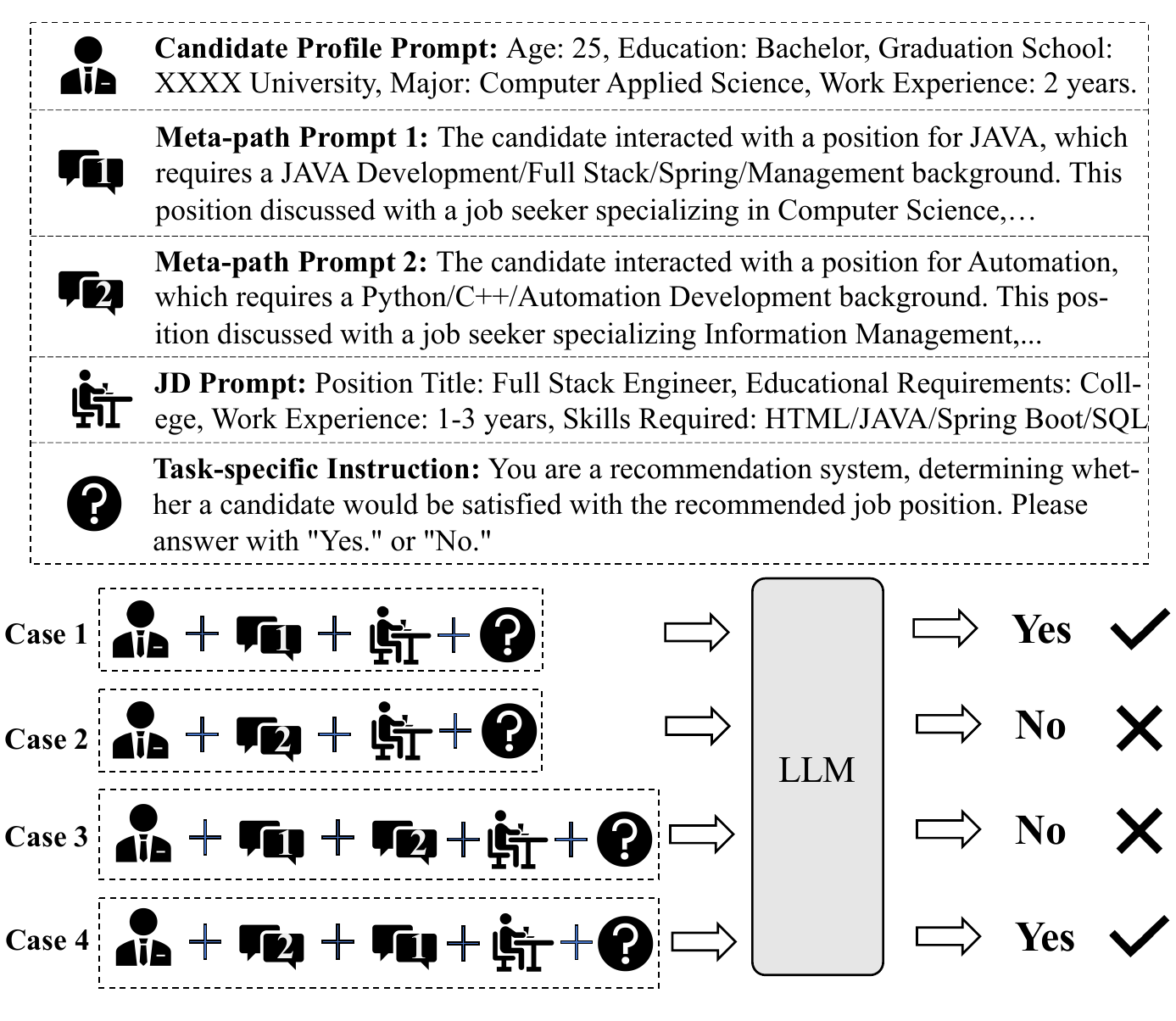}
    \caption{The real cases of path weight and position bias of meta-path prompt input for LLM.}
    \label{fig:path_case}
\end{figure}

The node information, i.e., the description of candidates or JD, then  will be filled in the meta-path prompt template to generate the final prompt data in our dataset. The real case can be referred to in Figure~\ref{fig:path_case}. In addition, to avoid too similar meta-paths leading to redundancy, we define a simple similarity metric as follows,
{\setlength\abovedisplayskip{0.05cm}
\setlength\belowdisplayskip{0.05cm}
\begin{equation}
\mathcal{S}_{i,j} = \frac{|P_i \cap P_j |}{|P_i \cup P_j|}, ~~~~P_i, P_j \in \Phi_P,
\label{eq:pathsim}
\end{equation}}
where $\Phi_P$ denotes the set of sampled paths for a candidate. $P_i, P_j$ indicates two meta-paths in $\Phi_P$. $|P_i \cap P_j|$ is the number of tokens that exist simultaneously in two paths, $P_i \cup P_j$ is the union of them. We ensure that $\mathcal{S}_{i, j} \leq \gamma$ between the final selected $M$ paths and $0 \leq \gamma \leq 1$ is a hyperparameter.

\subsubsection{Path Debiasing and Soft Selection}
Different from the traditional network embedding, sequence-based meta-path prompts would lead to two challenges for LLM to understand the candidates' behavior sub-graph.
\begin{myDef}
\textbf{Influence of Path Weight}. Different meta-paths would present different weights for the model decision.
\end{myDef}

\begin{myDef}
\textbf{Position Bias of Path Prompt}. The position bias of the order of path prompts brings unstable answers.
\end{myDef}

These two challenges appeared when recognizing the pre-trained large language model as a recommender, which hinders the effective modeling of semantic relationships in the graph by LLM recommendation models. To provide a more intuitive explanation, we extracted a real-world case from the log of a popular recruitment platform and visualized them in Figure~\ref{fig:path_case}. Specifically, for a job seeker in the IT industry, given his Candidate Profile Prompt, Meta-path Prompt 1, and Meta-path Prompt 2, we further feed the LLM with a Task-specific Instruction belonging to point-wise recommendation. The LLM recommender is expected to output the decision of ``Yes'' or ``No'' to present the preference of the candidate. Challenge 1 corresponds to Case 1 and Case 2 in this figure. We can find that the same profile and task description with different behavior meta-paths forces LLM to make different predictions. Obviously, the diversity of technology stacks in Path 1 reveals the candidate's preference for full-stack development, and compared to Path 2, the background of path-related job seeker is more close to our candidate. Therefore, for this candidate, Path 1 is evidently more important for the final decision. For Challenge 2, if we construct the input sequence as Case 3, i.e., the order is meta-path prompt 1 $\rightarrow$ meta-path prompt 2, the LLM outputs the wrong answer ``No''. But with a reverse path prompt order, the LLM is able to provide an accurate prediction. Similar to the widely known 
position bias of candidate items~\cite{wu2023survey}, the position of context prompt clearly misleads the model to generate unstable outputs.

To address the negative impact of these two challenges on the recommendation results, we carefully design an augmentation module specifically for the meta-path prompt, which consists of three concise but effective strategies. The first strategy is \textbf{Shuffle Mechanism}. When preparing domain data for the model's supervised fine-tuning (SFT), for each sample that contains multiple paths, we randomly shuffle the meta-path prompts in the sample $m$ times. Here $m$ denotes the conducted times of shuffling. This data augmentation technique allows the model to learn semantic invariance patterns from different combinations of paths, leading to more stable results. It enhances the robustness of the model without introducing redundant information. The second strategy is \textbf{Path Soft Selector}. In this work, we regard the path sampling process in Behavior Meta-path Prompt Generation as a hard selection to heuristic selects semantically rich paths. The Path Soft Selector is used to further adaptively assign a learned weight distribution to the constructed meta-path prompts. Firstly, for a given meta-path prompt $\mathcal{M}_i , i \in \{1, 2, ..., M\} $ ($M$ denotes the number of paths), we obtain the LLM word embedding $e_t$ of each token $t \in \mathcal{M}_i$. So, the meta-path embedding $H_i$ of $\mathcal{M}_i$ can be obtained via a mean pooling as follows,
\begin{equation}
H_i = \frac{1}{|\mathcal{M}_i|} \sum_{t \in \mathcal{M}_i} e_t, ~~~ i \in \{1, 2, ..., M\}.
\label{eq:pathemb}
\end{equation}
Then we propose a soft selector to calculate the weight for each meta-path embedding as: 
\begin{equation}
\alpha_{i} = \mathrm{softmax} (W_a H_i) = \frac{\mathrm{exp}(W_a H_i)}{\sum_{j=1}^{M} \mathrm{exp}(W_a H_j)},
\label{eq:selector}
\end{equation}
where $W_a \in \mathcal{R}^{1 \times d_e}$ is a trainable parameter, and $d_e$ denotes the dimension of $E_i$. To avoid the training collapse caused by changed value scale, we utilize a controller parameter $\lambda \in (0, 0.5]$ to update word embeddings in Eq. (\ref{eq:reslearn}).
\begin{equation}
\hat{e}_t = e_t + \lambda \cdot \alpha_i e_t, ~~~t \in \mathcal{M}_i,
\label{eq:reslearn}
\end{equation}
Compared with most existing tuned or non-tuned LLM models, our prompt augmentation mechanism considers phrase-based attention to distinguish different paths. Actually, this simple solution can be transferred to other similar situations, such as weighed sentence embeddings.

What's more, the third strategy is the \textbf{Hybrid Mechanism} which implements Shuffle Mechanism and Path Soft Selector simultaneously. This hybrid module is expected to address the both two challenges. We will evaluate these three strategies in the experiment section.

\subsection{LLM Instruction Tuning and Recommendation}
In this subsection, we will introduce the instruction tuning and recommendation process, which aims to align the used LLM with the recommendation task effectively and efficiently. For instruction tuning, we follow the general supervised fine-tuning method to minimize the autoregressive loss calculated by ground truth and corresponding LLM output. In our work, we mask the loss position of the prompt part. Specific prompt format, task-specific instruction, and ground truth have been introduced in the Preliminary section. However, direct fine-tuning of the entire model can be computationally intensive and time-consuming. To address this, we propose a lightweight fine-tuning strategy using LoRA, which involves freezing the pre-trained model parameters and introducing trainable rank decomposition matrices into each layer of the Transformer architecture. This approach facilitates lightweight fine-tuning while reducing GPU memory consumption. And the final learning objective can be computed as follows:
\begin{equation}
\mathcal{L}_f = \max _{\Theta_L} \sum_{(x, y) \in \mathcal{T}} \sum_{t = 1}^{|y|} \log \left(P_{\Theta+\Theta_L}\left(y_{t} \mid e_x, y_{<t}\right)\right),
\label{eq:loraloss}
\end{equation}
where $\Theta_L$ is the LoRA parameters and we only update LoRA parameters during the training process. Note that, different from existing fine-tuning frameworks for recommendation systems, we replace their token input $x$ by the embedding $e_x$ in Eq. (\ref{eq:loraloss}), since we update the prompt token embedding in the soft selector.

As for the recommendation process, since the trained model has learned the output format of our defined ground truth after several SFT alignment steps. So our designed answer parsing is a simple way. We catch the softmax probability of label generation (the token used to denote label, such as ``Yes./No.'' or ``[A]/[B]'' in our work ) in the position of model's output corresponding to that in the ground truth. Along this line, the final prediction probability is calculated.

\begin{table}[t]
	\caption{The statistics of datasets.}
	\centering
        \def\arraystretch{1.}
	\resizebox{.42\textwidth}{!}{
		\smallskip\begin{tabular}{c|cccccc}
            \toprule
			\textbf{Dataset}  & \textbf{\# Candidates} & \textbf{\# Jobs} & \textbf{\# Match} & \textbf{\# Interaction}\\
            \midrule
                RecrX & 12,440 & 19,318 & 23,879 & 54,147\\
			RecrY & 18,260 & 26,576 & 47,725 & 119,529\\
			\hline
		\end{tabular}
	}
	\label{table:Datasets}
\end{table}
\section{Experiments}

\subsection{Experimental Settings}
\subsubsection{Datasets.}
We conduct experiments on two datasets RecrX and RecrY with different scales which are collected from a real-world and large online recruitment platform in China to assess recommendation methods.
The datasets were constructed from the online logs and contained two kinds of behavior: Match and Interaction, corresponding to the matching set and interaction set mentioned in Problem Formulation. Besides, each candidate (and job) is associated with a descriptive text (i.e., resume or job description). The overall statistics are shown in Table~\ref{table:Datasets}. From the statistical data, it can be seen that job recommendation is a sparsely interactive scenario. The segmentation ratio of the training set and testing set is 5:1. Note that all sensitive or private information has been filtered out from the data.

\subsubsection{Baseline.} To provide a comprehensive evaluation of our GLRec model, we compare it against both LLM-based and related representative job recommendation methods. \textbf{RobertaRec}~\cite{liu2019roberta}: Candidate resume and JD text are encoded into fixed-length vectors using RoBERTa and then used to calculate similarity scores, enabling personalized recommendations. \textbf{HGT}~\cite{hu2020heterogeneous}: Heterogeneous Graph Transformer is a powerful graph learning model which propagates the embeddings (initialized by RoBERTa) of candidates and jobs on graph to capture high-order interactions. \textbf{DPGNN}~\cite{Yang_Hou_Song_Zhang_Wen_Zhao_2022}: The advanced job recommender Dual-Perspective GNN incorporates two different nodes for each candidate (or job) to model the two-way selection preference. \textbf{TALLrec}~\cite{DBLP:journals/corr/abs-2305-00447}: An advanced fine-tuned LLM recommender that uses instruction tuning on self-instruct data with users' historical interactions. The original backbone of its pre-trained model is LLaMA, and we change it by BELLE as the same as ours for the Chinese corpus.

\begin{table}[t]
    \centering
    \def\arraystretch{1.15}
        \caption{Job recommendation performance of AUC on test set, where $^*$ indicates the best result among baselines. Improve $\uparrow$ refers to the average enhancement achieved by GLRec in comparison to the baseline models. RX (RY) indicates RecrX (RecrY).}
        \resizebox{1.\columnwidth}{!}{
    \begin{tabular}{c|cc|cc|cc|cc}
        \toprule
        Task & \multicolumn{6}{c|}{Point-wise} & \multicolumn{2}{c}{Pair-wise} \\
        \midrule
        Split & \multicolumn{2}{c|}{Random} & \multicolumn{2}{c|}{OOD\_position} & \multicolumn{2}{c|}{OOD\_JD} & \multicolumn{2}{c}{Random} \\
        \midrule
        Dataset & RX & RY & RX & RY & RX & RY & RX & RY     \\
        \midrule
        RobertaRec & 0.710 & 0.734& 0.503 & 0.528 & 0.506 & 0.536 & 0.727 & 0.740    \\
        HGT & 0.744 & 0.756 & 0.572 & 0.595 & 0.576 & 0.593 & 0.747 & 0.751  \\
        DPGNN & 0.727 & 0.743 & 0.596 & 0.603 & 0.588 & 0.617 & 0.744 & 0.756  \\
        TALLrec & 0.842$^*$ & 0.829$^*$ & 0.770$^*$ & 0.788$^*$ & 0.766$^*$ & 0.798$^*$ & 0.849$^*$ & 0.825$^*$   \\
        \bf GLRec & \bf0.891 & \bf0.876 & \bf0.810 & \bf0.843 & \bf0.814 & \bf0.852 & \bf0.905 & \bf0.883    \\
                \midrule
                    Improve $\uparrow$ & 18.4\% & 14.1\% & 25.2\% & 28.3\% & 26.4\% & 29.8\% & 15.5\% & 13.2\%\\
        \bottomrule
        
    \end{tabular}
    
    }
    \label{tab:main-result}
    
\end{table}

\linespread{.94}
\begin{table*}[t]
\large
\setlength{\abovecaptionskip}{0.2cm}
\setlength{\belowcaptionskip}{0.2cm}
\caption{Some representative cases of our implemented models in the performance comparison experiment. Node 1 and Node 2 denote the nodes in a sampled meta-path of Candidate. RobRec denotes RobertaRec, and GT denotes Ground Truth.}
\centering
\resizebox{1.0\textwidth}{!}{
\begin{tabular}{m{4.5cm}|m{4.4cm}|m{5cm}|m{6.9cm}|c|ccc}
\toprule

Candidate & Node 1 (Job) & Node 2 (Job Seeker) & Target Job & GT & RobRec & TALLrec & GLRec \\
\midrule
Bachelor's degree, Computer Science, 3 years of work experience, skills... & Front-end Developer, Skill requirements: JavaScript / HTML5 /Vue & Bachelor's degree, Computer Applications Technology, Work experience: 2 years, skills... & Java, Qualification: Bachelor's degree, 5-10 years experience, Skill requirements: Java/System Architecture/Database & \bf{No}&\textcolor{blue}{No} &\textcolor{blue}{No}&\textcolor{blue}{No}\\
\midrule
Bachelor's degree, Business Administration, 9 years of work experience, skills... & Project Assistant, Skill requirements: Project Engineering Management & Bachelor's degree, International Economics and Trade, 3 years of work experience, skills... & Project Assistant, Qualification: Bachelor's degree, 3 years or more, Skill requirements: Project Engineering Management& \bf{Yes}&\textcolor{blue}{Yes} &\textcolor{blue}{Yes}&\textcolor{blue}{Yes}\\
\midrule
Bachelor's degree, Computer Applications Tech, 2 years of work experience, skills... & JAVA, Skill requirements: JAVA/Spring/Team Management Experience & Associate's degree, Internet of Things Technology, 4 years of work experience, skills... & Full Stack Engineer, Qualification: Associate's degree, 1-3 years of work experience, Skill requirements: JAVA/Spring/HTML& \bf{Yes}&\textcolor{red}{No} &\textcolor{blue}{Yes}&\textcolor{blue}{Yes}\\
\midrule
Bachelor's degree, Finance, 10 years of work experience, skills... & Functional Testing, Skill requirements: Software Testing/Requirement Alignment & Bachelor's degree, Financial Engineering, 2 years of work experience, skills... & Test Engineer, Qualification: Bachelor's degree, 3 years of work experience, Skill requirements: Functional Testing/Unit Testing& \bf{Yes}&\textcolor{red}{No} & \textcolor{red}{No} &\textcolor{blue}{Yes}\\

\bottomrule
\end{tabular}
}
\label{table:case}
\end{table*}
\subsubsection{Evaluation Metric.}
We evaluate the two tasks using the conventional metric: Area Under the Receiver Operating Characteristic (AUC), as our two tasks can be transferred to binary classification problems and the metric captures the similarity between our setting and predicting user interest in a target item. We do not employ ranking-based metrics because, during the fine-tuning process, the text sequence output of LLM requires ground truth for item order sequences, which, in reality, doesn't exist.

\subsubsection{Implementation Details.}
In this paper, we utilize BELLE-LLaMA-7B~\cite{BELLE} as the pre-trained LLM backbone due to its expanded Chinese vocabulary. The instruction-tuning and model inference, using LoRa, are conducted on 4 Tesla A100 80G GPUs. 
To ensure consistent sequence inputs within each batch (batch size is 32), we apply padding to sequences with a maximum length of 512. 
Our approach incorporates the meta-path prompt and user-specific task instructions as model inputs for personalized recommendations. In our experiments, we investigate the impact of different numbers of paths, specifically $M \in [0, 1, 2, 3]$, for GLRec, and the shuffled times $m=2$ for $M \geq 2$. In our work, we select paths with 3 nodes because they offer a balance between meaningful semantics and minimal redundancy with the experimental feedback.
Further details regarding the path prompt and instructions can be found in the Methodology section. 
Additionally, both RobertaRec and HGT have a token embedding dimension of 768, and HGT utilizes mean pooling to obtain the initial node embedding. 
For all methods, we optimize model parameters using the Adam~\cite{kingma2014adam} optimizer with a default learning rate of 1e-4, minimizing the MSE loss as the optimization objective. 
For the hyperparameters of update controller $\lambda$ and similarity threshold $\gamma$, we set $\lambda = 0.1$ and $\gamma = 0.3$ according to the experimental feedback. 
\subsection{Performance Comparison}
\subsubsection{Quantitative Comparison.} We conduct quantitative performance experiments on two datasets. As mentioned in the task definition in Section Methodology, the point-wise and pair-wise settings are implemented for evaluation. We also explore the influence of the OOD situation on different models. The experimental split settings of Random, OOD\_position, and OOD\_JD are introduced below:
\begin{itemize}
\item \textbf{Random}: We randomly split the training and testing dataset based on the interaction records of each user.
\item \textbf{OOD\_position}: The intersection on JD's ``job position'' feature between training set and testing set is empty.
\item \textbf{OOD\_JD}: The intersection on JD items between the training set and the testing set is empty.
\end{itemize}

\begin{figure}[t]
  \centering
  \subfigure[RecrX]{
    \label{vis:a} 
    \includegraphics[width=0.48\columnwidth]{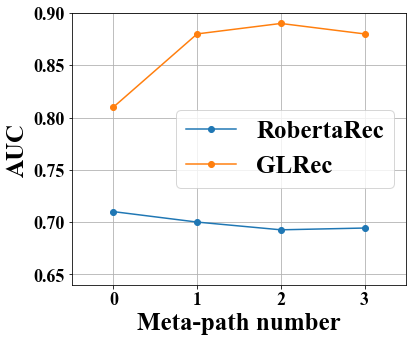}}
  \subfigure[RecrY]{
    \label{vis:b} 
    \includegraphics[width=0.48\columnwidth]{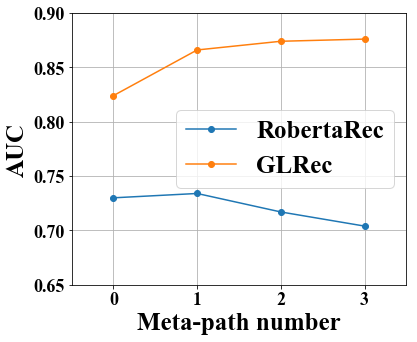}}
  \caption{The impact of meta-path number on model performance. }
  \label{fig:path_plot} 
\end{figure}
Our experimental results are reported in Table~\ref{tab:main-result}. Overall, our proposed GLRec model achieves the best performance among all baselines. There are distinctive score gaps between GLRec and all baselines according to the improvement in Table~\ref{tab:main-result}. It demonstrates the superiority and adaptability of the large-scale model framework that incorporates relationship understanding and extensive semantic knowledge in the job recommendation scenario. What's even more exciting is that GLRec demonstrates impressive performance on OOD tasks. While its performance may decline slightly compared to the random setting, our model achieves a significant breakthrough compared to other models, which essentially results in near-random guessing. This phenomenon illustrates the necessity of utilizing knowledge association for model generalization. Going deeper into the part of baselines, the graph-based HGT and DPGNN outperform the conventional dual-tower matching model (RobertaRec) in the context of job recommendation, which further proves the significance of learning relationships. What's more, we find that most models perform better on the pair-wise task than that of point-wise task. That is to say, directly determining whether an item is suitable is more challenging than comparing its priority with another item.
\subsubsection{Qualitative Comparison.}
To give a more intuitive visualization, some qualitative comparison results produced by models are shown in Table~\ref{table:case}, where the \textcolor{blue}{true} (\textcolor{red}{false}) prediction is highlighted in \textcolor{blue}{blue} (\textcolor{red}{red}) font. Specifically, the first two rows are straightforward, allowing multiple models to predict accurately. In the third row, solely using the user's profile isn't sufficient for prediction. It's crucial to note that the JAVA position (Node 1) the user interacted with aligns well with the target job in skill requirements. Consequently, only TALLrec and GLRec produced correct predictions. The final row emphasizes the significance of higher-order interactions, i.e., path, in LLM recommendations. Although there's a perceived mismatch between the candidate's finance major and the target job, interactions within the testing engineer and fintech sectors provide nuanced hints. For such complex cases, while the TALLrec model, relying on past behaviors, errs, only the GLRec model predicts correctly.



\subsection{The Impact of Meta-path Number}
We investigate the impact of meta-path number on the effectiveness of GLRec.
Here we evaluate the point-wise performance on Random setting using AUC for different numbers of meta-paths, ranging from 0 to 3.
We also input the meta-path prompt (removing extra instruction text for feature conciseness) into RobertaRec for comparison. From the line graph of Figure~\ref{fig:path_plot}, we can observe the following trends:
\begin{itemize}
\item For GLRec, the results consistently increase as the number of meta-paths increases. 
\item One notable observation is the significant improvement in GLRec's performance when transitioning from 0 meta-paths to 1 meta-path, and achieving the peak with only 2 or 3 meta-paths. The core increases from 0.71 to 0.88, indicating a substantial boost in recommendation effectiveness. This improvement suggests that the chain-of-thought ability of the LLM, inspired by in-context learning, plays a crucial role in GLRec's performance.
\item For RobertaRec, which does not incorporate behavior graph understanding, the values remain relatively stable across different meta-path numbers. The reason is that discriminative BERT-based model lacks the ability to effectively understand prompts like generative LLMs.
\end{itemize}
The results indicate that the inclusion of behavior graph understanding through meta-path prompt has a significant positive impact on the effectiveness of GLRec. By leveraging the rich information, GLRec gains a deeper understanding of user-item interactions, leading to improved performance, which provides sufficient evidence for graph effect.


\begin{figure}[t]
  \centering
  \subfigure[2 paths]{
    \label{vis:a} 
    \includegraphics[width=0.48\columnwidth]{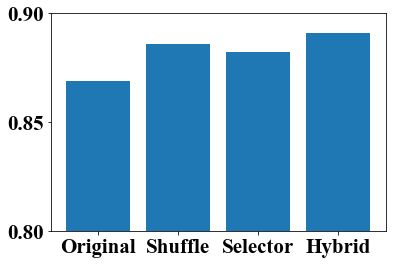}}
  \subfigure[3 paths]{
    \label{vis:b} 
    \includegraphics[width=0.48\columnwidth]{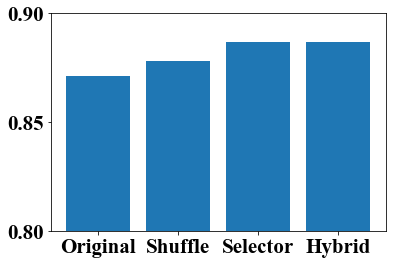}}
  \caption{The Impact of Bias of Meta-path Prompt. }
  \label{fig:path_bias} 
\end{figure}

\subsection{The Impact of Bias of Meta-path Prompt}
Due to the sequential nature of language model input, the construction of multi-path prompt sequences results in a human-induced position bias, or order bias, which disrupts the final decision-making of LLM model. Additionally, this input pattern does not allow the model to learn the importance of semantic information in different paths. Therefore, we design a path shuffle mechanism, a path soft selector, and a hybrid mechanism combining both to enhance the model's understanding of path information and mitigate bias. The experimental results on RecrX are reported in Figure~\ref{fig:path_bias}. Here the metric is AUC and the task is point-wise setting.

According to Figure~\ref{fig:path_bias}, our three strategies can all surpass the original input without path prompt augmentation in both two sub-experiments, which proves the necessity of path debiasing. Although the shuffle mechanism and soft selector have their own advantages and disadvantages in two different path scale experiments, both can relatively improve the quality of the results. And the hybrid module of both can bring more stable results, indicating that it is indeed necessary for the model to consider the position factors of input meta-paths and the influencing factors of different path prompts on decision-making in experiments, in order to cope with actual recommendation scenarios. Actually, in other similar scenarios, such as the input for LLM consists of multiple sentence prompts without prior order, our proposed shuffle mechanism and the soft selector can both play a certain role in enhancing the robustness of model training. We will continue to explore this property in our future work.

\section{Conclusion}
In conclusion, this paper introduced GLRec, a pioneering job recommendation model that seamlessly integrated large language models (LLMs) with behavior graph comprehension. By capitalizing on the semantic depth and vast knowledge inherent to LLMs, GLRec surpassed sufficient baselines in the quality of recommendations. The innovative meta-path prompt constructor effectively translated the intricate interaction details into natural language prompts, thereby refining personalized recommendation strategies. In the testing stage, rigorous experimental evaluations affirmed GLRec's efficacy, highlighting its dominant performance across real-world recruitment datasets. This investigation not only propelled the evolution of LLM-centric job recommendations but also charted fresh avenues for harnessing graph data in enhancing the personalized recommendation capabilities of LLMs.

\section*{Acknowledgments}
This research was partially supported by grants from National Key Research and Development Program of China (Grant No. 2021YFF0901003).

\bibliographystyle{aaai}
\bibliography{GLR}

\begin{thebibliography}{32}
\providecommand{\natexlab}[1]{#1}

\bibitem[{Bao et~al.(2023)Bao, Zhang, Zhang, Wang, Feng, and He}]{DBLP:journals/corr/abs-2305-00447}
Bao, K.; Zhang, J.; Zhang, Y.; Wang, W.; Feng, F.; and He, X. 2023.
\newblock TALLRec: An Effective and Efficient Tuning Framework to Align Large Language Model with Recommendation.
\newblock \emph{CoRR}, abs/2305.00447.

\bibitem[{Bian et~al.(2020)Bian, Chen, Zhao, Zhou, Hou, Song, Zhang, and Wen}]{bian2020learning}
Bian, S.; Chen, X.; Zhao, W.~X.; Zhou, K.; Hou, Y.; Song, Y.; Zhang, T.; and Wen, J.-R. 2020.
\newblock Learning to match jobs with resumes from sparse interaction data using multi-view co-teaching network.
\newblock In \emph{Proceedings of the 29th ACM International Conference on Information \& Knowledge Management}, 65--74.

\bibitem[{Cui et~al.(2022)Cui, Ma, Zhou, Zhou, and Yang}]{DBLP:journals/corr/abs-2205-08084}
Cui, Z.; Ma, J.; Zhou, C.; Zhou, J.; and Yang, H. 2022.
\newblock M6-Rec: Generative Pretrained Language Models are Open-Ended Recommender Systems.
\newblock \emph{CoRR}, abs/2205.08084.

\bibitem[{Dai et~al.(2022)Dai, Sun, Dong, Hao, Sui, and Wei}]{DBLP:journals/corr/abs-2212-10559}
Dai, D.; Sun, Y.; Dong, L.; Hao, Y.; Sui, Z.; and Wei, F. 2022.
\newblock Why Can {GPT} Learn In-Context? Language Models Secretly Perform Gradient Descent as Meta-Optimizers.
\newblock \emph{CoRR}, abs/2212.10559.

\bibitem[{Fu et~al.(2021)Fu, Liu, Zhu, Song, Zhang, and Wu}]{fu2021beyond}
Fu, B.; Liu, H.; Zhu, Y.; Song, Y.; Zhang, T.; and Wu, Z. 2021.
\newblock Beyond matching: Modeling two-sided multi-behavioral sequences for dynamic person-job fit.
\newblock In \emph{Database Systems for Advanced Applications: 26th International Conference, DASFAA 2021, Taipei, Taiwan, April 11--14, 2021, Proceedings, Part II 26}, 359--375. Springer.

\bibitem[{Hou et~al.(2023)Hou, Zhang, Lin, Lu, Xie, McAuley, and Zhao}]{DBLP:journals/corr/abs-2305-08845}
Hou, Y.; Zhang, J.; Lin, Z.; Lu, H.; Xie, R.; McAuley, J.~J.; and Zhao, W.~X. 2023.
\newblock Large Language Models are Zero-Shot Rankers for Recommender Systems.
\newblock \emph{CoRR}, abs/2305.08845.

\bibitem[{Hu et~al.(2021)Hu, Shen, Wallis, Allen-Zhu, Li, Wang, Wang, and Chen}]{hu2021lora}
Hu, E.~J.; Shen, Y.; Wallis, P.; Allen-Zhu, Z.; Li, Y.; Wang, S.; Wang, L.; and Chen, W. 2021.
\newblock Lora: Low-rank adaptation of large language models.
\newblock \emph{arXiv preprint arXiv:2106.09685}.

\bibitem[{Hu et~al.(2020)Hu, Dong, Wang, and Sun}]{hu2020heterogeneous}
Hu, Z.; Dong, Y.; Wang, K.; and Sun, Y. 2020.
\newblock Heterogeneous graph transformer.
\newblock In \emph{Proceedings of the web conference 2020}, 2704--2710.

\bibitem[{Ji et~al.(2023)Ji, Deng, Gong, Peng, Niu, Ma, and Li}]{BELLE}
Ji, Y.; Deng, Y.; Gong, Y.; Peng, Y.; Niu, Q.; Ma, B.; and Li, X. 2023.
\newblock BELLE: Be Everyone's Large Language model Engine.
\newblock \url{https://github.com/LianjiaTech/BELLE}.

\bibitem[{Jiang et~al.(2023)Jiang, Zhao, He, Wu, Zhang, and Fan}]{jiang2023knowledge}
Jiang, J.; Zhao, H.; He, M.; Wu, L.; Zhang, K.; and Fan, J. 2023.
\newblock Knowledge-Aware Cross-Semantic Alignment for Domain-Level Zero-Shot Recommendation.
\newblock In \emph{Proceedings of the 32nd ACM International Conference on Information and Knowledge Management}, 965--975.

\bibitem[{Kang et~al.(2023)Kang, Ni, Mehta, Sathiamoorthy, Hong, Chi, and Cheng}]{DBLP:journals/corr/abs-2305-06474}
Kang, W.; Ni, J.; Mehta, N.; Sathiamoorthy, M.; Hong, L.; Chi, E.~H.; and Cheng, D.~Z. 2023.
\newblock Do LLMs Understand User Preferences? Evaluating LLMs On User Rating Prediction.
\newblock \emph{CoRR}, abs/2305.06474.

\bibitem[{Kenthapadi, Le, and Venkataraman(2017)}]{Kenthapadi_Le_Venkataraman_2017}
Kenthapadi, K.; Le, B.; and Venkataraman, G. 2017.
\newblock Personalized Job Recommendation System at LinkedIn: Practical Challenges and Lessons Learned.
\newblock In \emph{Proceedings of the Eleventh ACM Conference on Recommender Systems}.

\bibitem[{Kingma and Ba(2014)}]{kingma2014adam}
Kingma, D.~P.; and Ba, J. 2014.
\newblock Adam: A method for stochastic optimization.
\newblock \emph{arXiv preprint arXiv:1412.6980}.

\bibitem[{Le et~al.(2019)Le, Hu, Song, Zhang, Zhao, and Yan}]{le2019towards}
Le, R.; Hu, W.; Song, Y.; Zhang, T.; Zhao, D.; and Yan, R. 2019.
\newblock Towards effective and interpretable person-job fitting.
\newblock In \emph{Proceedings of the 28th ACM International Conference on Information and Knowledge Management}, 1883--1892.

\bibitem[{Liu et~al.(2019)Liu, Ott, Goyal, Du, Joshi, Chen, Levy, Lewis, Zettlemoyer, and Stoyanov}]{liu2019roberta}
Liu, Y.; Ott, M.; Goyal, N.; Du, J.; Joshi, M.; Chen, D.; Levy, O.; Lewis, M.; Zettlemoyer, L.; and Stoyanov, V. 2019.
\newblock Roberta: A robustly optimized bert pretraining approach.
\newblock \emph{arXiv preprint arXiv:1907.11692}.

\bibitem[{Lu, El~Helou, and Gillet(2013)}]{Lu_El_Helou_Gillet_2013}
Lu, Y.; El~Helou, S.; and Gillet, D. 2013.
\newblock A recommender system for job seeking and recruiting website.
\newblock In \emph{Proceedings of the 22nd International Conference on World Wide Web}.

\bibitem[{Luo et~al.(2019)Luo, Zhang, Wen, and Zhang}]{luo2019resumegan}
Luo, Y.; Zhang, H.; Wen, Y.; and Zhang, X. 2019.
\newblock Resumegan: An optimized deep representation learning framework for talent-job fit via adversarial learning.
\newblock In \emph{Proceedings of the 28th ACM international conference on information and knowledge management}, 1101--1110.

\bibitem[{Penha and Hauff(2020)}]{penha2020does}
Penha, G.; and Hauff, C. 2020.
\newblock What does {BERT} know about books, movies and music? Probing {BERT} for Conversational Recommendation.
\newblock In \emph{RecSys}, 388--397. {ACM}.

\bibitem[{Qin et~al.(2018)Qin, Zhu, Xu, Zhu, Jiang, Chen, and Xiong}]{qin2018enhancing}
Qin, C.; Zhu, H.; Xu, T.; Zhu, C.; Jiang, L.; Chen, E.; and Xiong, H. 2018.
\newblock Enhancing person-job fit for talent recruitment: An ability-aware neural network approach.
\newblock In \emph{The 41st international ACM SIGIR conference on research \& development in information retrieval}, 25--34.

\bibitem[{Qiu et~al.(2021)Qiu, Wu, Gao, and Fan}]{qiu2021u}
Qiu, Z.; Wu, X.; Gao, J.; and Fan, W. 2021.
\newblock {U-BERT:} Pre-training User Representations for Improved Recommendation.
\newblock In \emph{{AAAI}}, 4320--4327. {AAAI} Press.

\bibitem[{Ramanath et~al.(2018)Ramanath, Inan, Polatkan, Hu, Guo, Ozcaglar, Wu, Kenthapadi, and Geyik}]{ramanath2018towards}
Ramanath, R.; Inan, H.; Polatkan, G.; Hu, B.; Guo, Q.; Ozcaglar, C.; Wu, X.; Kenthapadi, K.; and Geyik, S.~C. 2018.
\newblock Towards deep and representation learning for talent search at linkedin.
\newblock In \emph{Proceedings of the 27th ACM International Conference on Information and Knowledge Management}, 2253--2261.

\bibitem[{Shen et~al.(2018)Shen, Zhu, Zhu, Xu, Ma, and Xiong}]{shen2018joint}
Shen, D.; Zhu, H.; Zhu, C.; Xu, T.; Ma, C.; and Xiong, H. 2018.
\newblock A joint learning approach to intelligent job interview assessment.
\newblock In \emph{IJCAI}, volume~18, 3542--3548.

\bibitem[{Wu et~al.(2021{\natexlab{a}})Wu, Wu, Yu, Qi, Huang, and Xie}]{wu2021userbert}
Wu, C.; Wu, F.; Yu, Y.; Qi, T.; Huang, Y.; and Xie, X. 2021{\natexlab{a}}.
\newblock Userbert: Contrastive user model pre-training.
\newblock \emph{arXiv preprint arXiv:2109.01274}.

\bibitem[{Wu et~al.(2021{\natexlab{b}})Wu, Li, Zhao, Liu, Wang, Zhang, and Chen}]{wu2021learning}
Wu, L.; Li, Z.; Zhao, H.; Liu, Q.; Wang, J.; Zhang, M.; and Chen, E. 2021{\natexlab{b}}.
\newblock Learning the implicit semantic representation on graph-structured data.
\newblock In \emph{Database Systems for Advanced Applications: 26th International Conference, DASFAA 2021, Taipei, Taiwan, April 11--14, 2021, Proceedings, Part I 26}, 3--19. Springer.

\bibitem[{Wu et~al.(2023)Wu, Zheng, Qiu, Wang, Gu, Shen, Qin, Zhu, Zhu, Liu et~al.}]{wu2023survey}
Wu, L.; Zheng, Z.; Qiu, Z.; Wang, H.; Gu, H.; Shen, T.; Qin, C.; Zhu, C.; Zhu, H.; Liu, Q.; et~al. 2023.
\newblock A Survey on Large Language Models for Recommendation.
\newblock \emph{arXiv preprint arXiv:2305.19860}.

\bibitem[{Yan et~al.(2019)Yan, Le, Song, Zhang, Zhang, and Zhao}]{Yan_Le_Song_Zhang_Zhang_Zhao_2019}
Yan, R.; Le, R.; Song, Y.; Zhang, T.; Zhang, X.; and Zhao, D. 2019.
\newblock Interview Choice Reveals Your Preference on the Market: To Improve Job-Resume Matching through Profiling Memories.
\newblock In \emph{Proceedings of the 25th ACM SIGKDD International Conference on Knowledge Discovery {\&} Data Mining}.

\bibitem[{Yang et~al.(2022)Yang, Hou, Song, Zhang, Wen, and Zhao}]{Yang_Hou_Song_Zhang_Wen_Zhao_2022}
Yang, C.; Hou, Y.; Song, Y.; Zhang, T.; Wen, J.-R.; and Zhao, W.~X. 2022.
\newblock Modeling Two-Way Selection Preference for Person-Job Fit.
\newblock In \emph{Sixteenth ACM Conference on Recommender Systems}.

\bibitem[{Yu et~al.(2023)Yu, Liu, Wu, Yu, Yu, and Zhang}]{yu2023untargeted}
Yu, Y.; Liu, Q.; Wu, L.; Yu, R.; Yu, S.~L.; and Zhang, Z. 2023.
\newblock Untargeted attack against federated recommendation systems via poisonous item embeddings and the defense.
\newblock In \emph{Proceedings of the AAAI Conference on Artificial Intelligence}, volume~37, 4854--4863.

\bibitem[{Zhang et~al.(2023)Zhang, Xie, Hou, Zhao, Lin, and Wen}]{DBLP:journals/corr/abs-2305-07001}
Zhang, J.; Xie, R.; Hou, Y.; Zhao, W.~X.; Lin, L.; and Wen, J. 2023.
\newblock Recommendation as Instruction Following: {A} Large Language Model Empowered Recommendation Approach.
\newblock \emph{CoRR}, abs/2305.07001.

\bibitem[{Zhang and Wang(2023)}]{zhang2023prompt}
Zhang, Z.; and Wang, B. 2023.
\newblock Prompt Learning for News Recommendation.
\newblock \emph{arXiv preprint arXiv:2304.05263}.

\bibitem[{Zhao et~al.(2023)Zhao, Zhao, Li, He, Wang, and Fan}]{zhao2023cross}
Zhao, C.; Zhao, H.; Li, X.; He, M.; Wang, J.; and Fan, J. 2023.
\newblock Cross-Domain Recommendation via Progressive Structural Alignment.
\newblock \emph{IEEE Transactions on Knowledge and Data Engineering}.

\bibitem[{Zheng et~al.(2023)Zheng, Qiu, Hu, Wu, Zhu, and Xiong}]{zheng2023generative}
Zheng, Z.; Qiu, Z.; Hu, X.; Wu, L.; Zhu, H.; and Xiong, H. 2023.
\newblock Generative Job Recommendations with Large Language Model.
\newblock arXiv:2307.02157.

\end{thebibliography}

\end{document}